\documentclass[letterpaper, 10 pt, conference]{ieeeconf}  

\IEEEoverridecommandlockouts                              

\usepackage{amsmath} 
\usepackage{graphicx} 
\usepackage{array}
\usepackage{booktabs}
\usepackage{amsfonts}
\usepackage{bm}
\usepackage{physics}
\usepackage{url}
\usepackage{hyperref}

\title{\LARGE \bf
DiSA-IQL: Offline Reinforcement Learning for Robust Soft Robot Control under Distribution Shifts}
\author{Linjin He$^{1}$, Xinda Qi$^{2}$, Dong Chen$^{3}$, Zhaojian Li$^{4}$, Xiaobo Tan$^{2}$
\thanks{$^{1}$Linjin He is with the Department of Data Science and Analysis, Georgetown University, Washington, DC, USA. Email: {\tt\small lh1085@georgetown.edu}}%
\thanks{$^{2}$Xinda Qi and Xiaobo Tan are with the Department of Electrical and Computer Engineering, Michigan State University,
Lansing, MI, USA. Emails: {\tt\small qixinda@msu.edu; xbtan@egr.msu.edu}}%
\thanks{$^{3}$Dong Chen is with the Department of Agricultural \& Biological Engineering, Mississippi State University, Starkville, MS, USA. Email: {\tt\small dc2528@msstate.edu}}%
\thanks{$^{4}$Zhaojian Li is with the Department of Mechanical Engineering,  Michigan State University, Lansing, MI, USA. Email: {\tt\small lizhaoj1@egr.msu.edu}}%
}

\begin{document}

\maketitle
\thispagestyle{empty}
\pagestyle{empty}

\begin{abstract}
Soft snake robots offer remarkable flexibility and adaptability in complex environments, yet their control remains challenging due to highly nonlinear dynamics. Existing model-based and bio-inspired controllers rely on simplified assumptions that limit their performance. Deep reinforcement learning (DRL) has recently emerged as a promising alternative, but online training is often impractical because of costly and potentially damaging real-world interactions. Offline RL provides a safer option by leveraging pre-collected datasets, but it suffers from distribution shift, which degrades generalization to unseen scenarios. To overcome this challenge, we propose DiSA-IQL (Distribution-Shift-Aware Implicit Q-Learning), an extension of IQL that incorporates robustness modulation by penalizing unreliable state–action pairs to mitigate distribution shift. We evaluate DiSA-IQL on goal-reaching tasks across two settings: in-distribution and out-of-distribution evaluation. Simulation results show that DiSA-IQL consistently outperforms baseline models, including Behavior Cloning (BC), Conservative Q-Learning (CQL), and vanilla IQL, achieving higher success rates, smoother trajectories, and greater robustness.
\end{abstract}


\section{INTRODUCTION}
Soft robots have attracted considerable research attention across various applications, such as fruit harvesting \cite{wang2023development}, medical surgery \cite{cianchetti2018biomedical}, and search-and-rescue operations \cite{wen2024design}. Among them, soft snake robots have gained particular interest due to their unique locomotion modes, high flexibility, and adaptability to complex and cluttered environments  \cite{8404900, Fu_2020}. However, the control of such robots remains challenging due to their highly non-linear dynamics, large state-space dimensions, and limited or sparse feedback \cite{liu2025integrating}.

Traditional control methods for soft robots rely on mathematical modeling of the robot’s kinematics and dynamics, often using simplified geometric or reduced-order models (e.g., piecewise constant curvature) to reduce complexity. However, these methods are highly sensitive to modeling errors and can be computationally intensive \cite{6225163,5649110}. Additionally, bio-inspired approaches, such as serpenoid curve–based controllers and gait-based methods derived from real snake locomotion, generate sinusoidal or wave-like patterns to achieve forward or lateral movement \cite{branyan2017soft, 1521742}. While easier to implement and effective on flat terrain, these methods depend on simplified environmental assumptions and often lack robustness in complex or uncertain environments.

Recently, reinforcement learning (RL) has emerged as a promising alternative for controlling soft robots. Unlike traditional model-based or bio-inspired methods, RL does not require explicit kinematic or dynamical models; instead, it learns control policies directly from interaction with the environment. This allows the robot to adapt to nonlinear dynamics, unmodeled friction, and uncertain terrains, thereby improving robustness and generalization in real-world scenarios \cite{graule2022somogym, jitosho2023reinforcement, chen2024data, Qi2024BER}. For instance, Graule et al. \cite{graule2022somogym} introduced SoMoGym, a soft robot training library, where multiple RL algorithms were benchmarked on tasks such as planar block pushing, snake locomotion, and in-hand manipulation, demonstrating the potential of RL for effective soft robot control. In our recent study \cite{Qi2024BER}, we proposed Back-Stepping Experience Replay (BER), an augmentation strategy that generates reversed transitions to improve exploration efficiency. Applied to a soft snake robot, BER consistently outperforms state-of-the-art baselines, achieving a higher success rate and significantly faster average locomotion speed. Despite these achievements, current online RL approaches often require extensive training interactions, making them sample inefficient and difficult to deploy directly on physical soft robots \cite{sinha2022s4rl, robotics8010004}.

One promising direction is offline RL, which reduces reliance on real-time exploration by leveraging pre-collected datasets \cite{levine2020offlinereinforcementlearningtutorial}. Beyond data efficiency, offline RL improves safety by avoiding excessive hardware wear, facilitates scalable policy training, and supports sim-to-real transfer by integrating simulated and real data \cite{levine2020offlinereinforcementlearningtutorial, tiboni2023dropo}. Recent studies have applied goal-conditioned offline RL to deformable objects \cite{laezza2024offline} and combined learned dynamics models with RL in soft robot simulators \cite{berdica2024reinforcement}. However, applications to soft robots, and particularly to soft snake robots, remain largely unexplored. Moreover, offline RL faces challenges such as distribution shift, where the learned policy may exploit out-of-distribution actions not adequately represented in the training dataset, leading to suboptimal or unsafe behaviors \cite{pmlr-v97-fujimoto19a, NEURIPS2020_0d2b2061, kostrikov2021offlinereinforcementlearningimplicit}. To address these challenges, we propose a distribution shift–aware offline RL framework specifically designed for soft snake robots.  First, we systematically evaluate and benchmark multiple classical offline RL algorithms, i.e., Behavior Cloning (BC), Conservative Q-Learning (CQL), and Implicit Q-Learning (IQL), on the locomotion and navigation tasks of a soft snake robot. Furthermore, we enhance IQL with an out-of-distribution action suppression mechanism to mitigate distribution shift in soft robot control.

\section{Problem formulation}

\subsection{Soft snake robot and serpentine locomotion}
The modular soft snake robot consists of multiple 2D soft pneumatic bending actuators, enabling a planar traveling-wave deformation along its body (Fig. 1a). A bio-inspired serpentine locomotion is generated with the help of the friction anisotropic snake skins when the traveling-wave is activated. The actuation of the soft bending actuator is applied by using the pressure differences between the local air chambers on the two sides of the snake body. 

As in our previous work \cite{Qi2024BER}, the shape of the snake robot is modeled as an inextensible curve in 2D space. The position of each point of the backbone curve of the robot at time $t$ is:
\begin{equation}
    \bm{X}(s,t) = \big(X(s,t), Y(s,t)\big), \quad s \in [0,L]
\end{equation}
where $s$ is the curve length from the tail of the robot.

The center of mass (COM) $\overline{\bm{X}}(t)$ and the main direction $\overline{\theta}$ of the curved snake robot (Fig. 1b) are derived from the averaged local position and orientation of each point of the snake robot \cite{Qi2024BER}. Then, the shape of the snake robot can be expressed using the mean-zero anti-derivative $I_0$ \cite{hu2012slithering} ($I_0[f](s,t)$ $= \int_0^s f(s',t)\dd{s}' -\frac{1}{L}\int_0^L\dd{s}\int_0^s\dd{s'}f(s',t) $):
\begin{equation}
    \bm{X}(s, t) = \overline{\bm{X}}(t) + I_0[\bm{X}_s](s,t)
\end{equation}
\begin{equation}
    \theta(s, t) = \overline{\theta}(t) + I_0[\kappa](s,t)
\end{equation}
where $\theta$ is the local orientation (angle between the local tangent direction and $X$ axis), $\bm{X}_s = (\cos{\theta}, \sin{\theta})$, and $\kappa(s,t)$ is the local curvature. The local curvature is determined by the pressure difference on the two sides:
\begin{equation}
    \kappa(s, t) = K_b \cdot \Delta p(s,t)
\end{equation}
where $K_b$ is an elasticity constant and $\Delta p(s,t)$ is the local pressure difference at $s$, which is computed from the pressures in different air channels. The pressure in different channels (Fig. 1a) is set as: 
\begin{equation}
    {
    \begin{split}
        p_i = p_m\sin{( c \cdot \frac{2\pi}{T} t_r + \frac{(i-1)\cdot\pi}{2})} + b_{i, pre} \\+  (b_i - b_{i, pre}) \frac{t_r}{T}
    \end{split}
    }
\end{equation}
where $t_r \in [0,T]$ and $T$ is the actuation period. $p_m$ and $b_i$ are the pressure magnitude and bias of $i$-th channel, respectively, $i \in \{1,2,3,4\}$. $b_{i, pre}$ is a one step history of $b_i$, and $c \in \{1, -1\}$ controls propagation direction of the traveling-wave.

Based on the shape model of the snake robot, the dynamic model is then built by analyzing the force interaction, which includes the complex friction between the snake skin and the ground \cite{Qi2024BER}. The anisotropic friction $\bm{f}_{fric}$ is described by:
\begin{equation}
\begin{cases}
    \bm{f}_\text{fric} = -\rho g (\mu_t(\hat{\bm{u}}\cdot\hat{\bm{t}})\hat{\bm{t}} + \mu_l(\hat{\bm{u}} \cdot \hat{\bm{f}}) \hat{\bm{f}} ) \\
    \mu_l = \mu_f H(\hat{\bm{u}}\cdot\hat{\bm{f}}) + \mu_b(1-H(\hat{\bm{u}}\cdot\hat{\bm{f}}))
\end{cases}
\end{equation}
where $\hat{\bm{u}}$ denotes the unit local velocity, $\mu_f$, $\mu_b$, and $\mu_t$ represent the friction coefficients in different local directions: forward along the body $\hat{\bm{f}}$, backward $\hat{\bm{b}}$, and transverse $\hat{\bm{t}}$, respectively.
The dynamic of each point is determined by:
\begin{equation}
    \rho \ddot{X}(s,t) = f_{\text{fric}}(s,t) + f_{\text{int}}(s,t)
\end{equation}
where $\rho$ denotes density, and $f_\text{int}$ denotes the internal force in the robot body, which could be eliminated during the dynamic computation \cite{Qi2024BER} \cite{hu2012slithering}.

\begin{figure}
    \centering
    \includegraphics[width=0.95\linewidth]{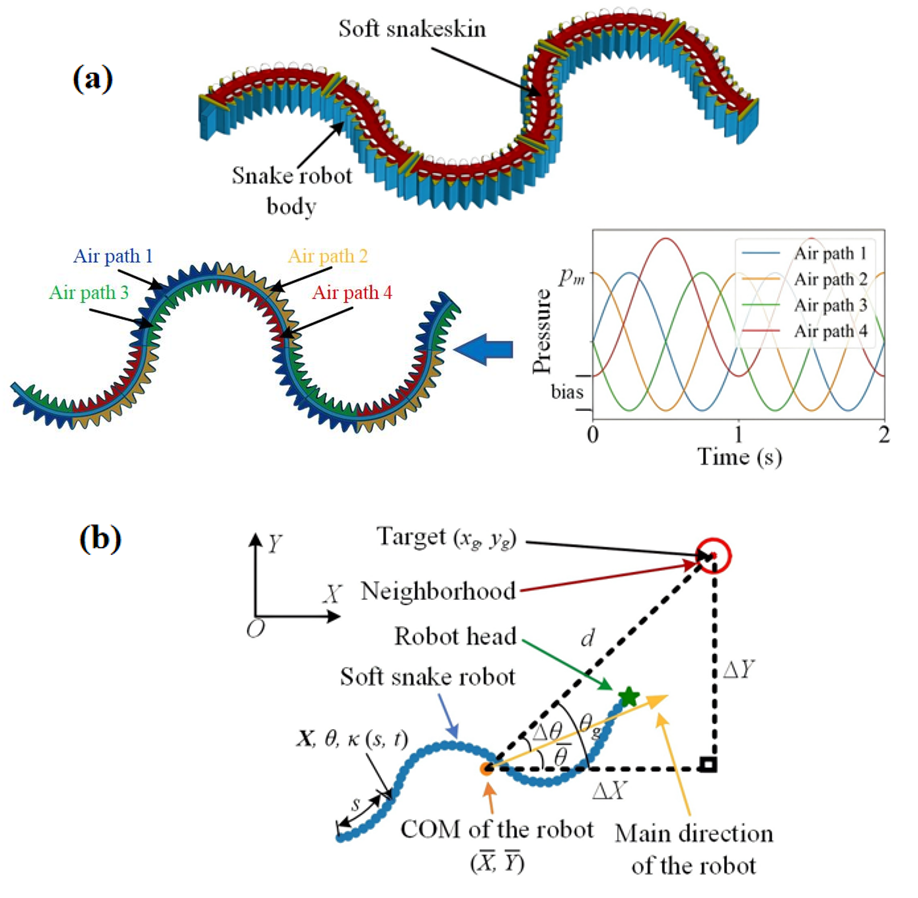}
    \caption{(a) Soft snake robot and its pneumatic air paths for actuation. (b) Target-reaching task of the snake robot, where the RL agent drives the robot’s center of mass (COM) from the start point into the neighborhood of the target point.}
    \label{fig:placeholder8}
    \vspace{-15pt}
\end{figure}

\subsection{Task definition and RL formulation}
A target-reaching task (Fig. 1b) is studied for the control of the soft snake robot by utilizing its serpentine locomotion. Specifically, an RL agent drives the snake robot from the original start point to a random target point, where the COM of the robot represents its position. The success criteria are set as whether the robot reaches a small neighborhood around the target point:
\begin{equation}
    \text{Success:} \quad \text{if } \| \overline{\bm{X}} - \bm{X}_{g} \| \leq \epsilon
\end{equation}

A Markov Decision Process (MDP), $\mathcal{M} = (S, A, P, r, \gamma)$, is formulated for the task of the robot:
\begin{itemize}
    \item \textbf{State space $S$}: 
    The state (which is also the observation) at time step $t$ is defined as:
    \begin{equation}
        s_t = [\Delta x_t, \; \Delta y_t, \; \Delta \theta_t, \; b_{i,t}, \; b_{i, pre, t}]
    \end{equation}
    where $\Delta x_t = x_g - x_t$ and $\Delta y_t = y_g - y_t$ are the relative distances from the robot head to the target in $X$ and $Y$ direction, respectively. 
    $\Delta \theta_t = \theta_g - \overline{\theta}$ is the deflection, where $\theta_g = \arctan(\Delta x_t, \Delta y_t)$. $b_{i,t}$, and $b_{i, pre, t}$ are the pressure bias in $i$-th channel and its one step history. 
    
    \item \textbf{Action space $A$}: 
    The control action at each step is defined as:
    \begin{equation}
        a_t = (b_{i,t}, c_t)
    \end{equation}
    Here, $b_{i,t}\in[0,1]$ denotes the pressure bias of the $i$-th pneumatic channel, which shifts the baseline pressure level in the traveling-wave actuation, and $c_t\in\{-1,1\}$ controls the propagation direction of the traveling wave(forward or backward). The pressure-related action variables are continuous, while $c_t$ remains a binary variable that selects the wave propagation direction.

    \item \textbf{Transition dynamics}:
    A model-free approach is used in the learning of the RL agent. Thus, there is no explicit model information needed in the learning process.

    \item \textbf{Reward function $r(s_t,a_t)$}: 
    The reward function is designed as a combination of a sparse position reward (success) and the punishment for the distance and the deflection of the robot towards the target:
    \begin{equation}
        r_t = - \Big( \alpha \frac{d_t}{d_0} + \beta \frac{|\Delta \theta_t|}{\pi} \Big) + \mathbb{I}\{\text{goal reached}\} \cdot R_{\text{success}}
    \end{equation}
    where $d_t$ is the current distance to the target, $d_0$ is the initial distance, $\Delta \theta_t$ is the current deflection. For notational simplicity, we use $r_t$ as a shorthand for $r(s_t,a_t)$. $\alpha,\beta$ are weighting coefficients, which are 0.15 and 1 in the RL training, respectively.
    A large terminal reward $R_{\text{success}}$ is given once the robot head reaches within the target neighborhood. The heading deflection term may become sensitive near the target; however, its impact is limited by setting a large final reward to terminate the training process and by normalizing the term to keep it within a certain range.

    \item \textbf{Discount factor $\gamma$}:
    A discount factor $\gamma \in (0,1)$ is used to balance immediate and long-term rewards.
\end{itemize}

Then, an RL agent is trained to learn how to drive the robot to reach the target. The objective of the RL agent is to learn an optimal policy $\pi^*$ that maximizes the expected discounted return:
\begin{equation}
    \pi^* = \arg \max_\pi \mathbb{E}_\pi \left[ \sum_{t=0}^T \gamma^t r(s_t,a_t) \right]
\end{equation}

\section{Distribution-Shift-Aware Implicit Q-Learning Framework}
\subsection{Offline reinforcement learning}
By utilizing a static dataset of trajectories, offline RL is able to train effective policies without the need to interact with the environments, which saves the data collection cost from new simulations or real experiments \cite{levine2020offlinereinforcementlearningtutorial}. The dataset includes state-action pairs from different sources, including demonstration, behavior from other policies, or a mixture of expert behavior. 

Behavior cloning (BC) is a simple and widely used imitation-based offline learning method among the imitation-based methods \cite{levine2020offlinereinforcementlearningtutorial}. It reduces the offline RL problem into a supervised learning task, avoiding reasoning about the returns or Q-values at the cost of limited generalization ability. With a dataset $\mathcal{D} = \{(s, a)\}$, the objective is to learn a policy $\pi^*$ that imitates the action distribution of the dataset:
\begin{equation}
    \pi^* = \text{arg} \max_{\pi_\theta} \mathbb{E}_{(s,a)\sim\mathcal{D}}[\log\pi_{\theta}(a|s)]
\end{equation}

Conservative Q-Learning (CQL) is another method of learning policies from a static dataset by using a pessimistic Q-function \cite{NEURIPS2020_0d2b2061}. Instead of trusting all Q-values trained by using the standard Bellman equation in Q-Learning, CQL explicitly penalizes the Q-values of the out-of-distribution (OOD) actions. Instead of directly bootstrapping within the dataset, CQL adds a regularizer that decreases the Q-values for actions not in the dataset, making the Q-learning pessimistic about unseen actions and avoiding overestimation. The loss function of the CQL is built based on that of the Q-Learning (QL):
\begin{equation}
\begin{split}
    \mathcal{L}&_{CQL}(Q) = \mathcal{L}_{QL}(Q) \\ &+ \alpha(\mathbb{E}_{s \sim \mathcal{D}, a\sim\pi(\cdot|s)} [Q(s,a) ] - \mathbb{E}_{(s,a) \sim \mathcal{D}}[Q(s,a)]  )
\end{split}
\end{equation}

Implicit Q-Learning (IQL) avoids both explicit policy imitation and conservative penalty \cite{kostrikov2021offlinereinforcementlearningimplicit}. It utilizes expectile regression and advantage weighted regression to bias the policy towards the high-return trajectories, making the algorithm stable and also avoiding overestimation of the OOD actions. The expectile regression provides a conservative estimation of the value function:
\begin{equation}
    V(s) = \text{arg} \min_v \mathbb{E}_{(s,a,r,s')\sim \mathcal{D}} [L^\tau(Q(s,a)-v(s))]
\end{equation}
where $L^\tau(x) = |\tau - \bm{1}_{x<0}|x^2$, $\tau < 0.5$ biases the value function $V(s)$ to the underestimated returns.

Then, by using the advantage function $A(s, a) = Q(s, a) - V(s)$, the policy is trained to favor the actions with higher advantages:
\begin{equation}
    \pi^* = \arg \max_{\pi_\theta} \mathbb{E}_{(s,a)\sim\mathcal{D}}[e^{A(s,a)/\beta} \log \pi_\theta(a|s)]
\end{equation}
where $\beta$ is a temperature parameter controlling the preference for high-advantage actions.

\subsection{Distribution-shift-aware implicit learning (DiSA-IQL)}

In our initial tests of soft robot control, IQL demonstrated
competitive performance by implicitly weighting actions
present in the dataset and thereby avoiding explicit pessimism.
However, standard IQL is highly susceptible to performance
degradation under distribution shift, because it does not explicitly penalize out-of-distribution (OOD) actions. As a result, the learned policy may exploit inflated Q-values and fail to generalize.

To address this issue, we propose a distribution-shift-aware
implicit Q-learning framework (DiSA-IQL), which introduces
a robustness penalty to suppress unreliable state–action pairs, inspired by distributionally robust RL, where uncertainty in the dataset distribution is handled through conservative value estimation
\cite{Derman2021RobustRL, panaganti2023bridgingdistributionallyrobustlearning}.
Similar to IQL, the value function $V_\psi$ is learned using
expectile regression:
\begin{equation}
L_V(\psi) =
\mathbb{E}_{(s,a)\sim \mathcal{D}}
\Big[
L^\tau\!\left(Q_\theta(s,a)-V_\psi(s)\right)
\Big],
\end{equation}
where $\tau\in(0,1)$ is the expectile parameter and
$\mathcal{D}$ denotes the offline dataset.
The Q-function is then trained using a robust Bellman
regression \cite{panaganti2023bridgingdistributionallyrobustlearning}:
\begin{equation}
\begin{split}
L_Q(\theta) 
&= \mathbb{E}_{(s,a,r,s') \sim \mathcal{D}} \\
&\quad \left[ \left( Q_\theta(s,a) - 
\Big( r + \gamma (1-d)\,( V_\psi(s') - P(s,a) ) \Big) \right)^2 \right],
\end{split}
\end{equation}
where $P(s,a)$ is a robustness penalty that discourages
state–action pairs with high uncertainty. Intuitively,
actions that are rarely observed in the dataset may have
unreliable value estimates due to extrapolation errors.
Introducing the penalty $P(s,a)$ effectively reduces the
Bellman target for such transitions, preventing the policy
from exploiting overestimated Q-values.

Following the framework of distributionally robust learning, the penalty can be derived based on different uncertainties
sets \cite{Derman2021RobustRL, Kumar2020CQL,
panaganti2023bridgingdistributionallyrobustlearning},
including Total Variation (TV), Wasserstein,
Kullback--Leibler (KL), and Chi-square divergences.
In practice, we estimate uncertainty from the visitation counts of state–action pairs.

Let $N(s,a)$ denote the visitation count of $(s,a)$ in the
dataset. The penalty term is defined as:
\begin{equation}
P(s,a)=
\alpha \,
\min\!\left(
\frac{\log N(s,a)}{N(s,a)},
\log |\mathcal{S}|
\right),
\end{equation}
where $\alpha>0$ controls the strength of the robustness
regularization and $|\mathcal{S}|$ is the cardinality of the
state space. This formulation assigns larger penalties to
rarely visited state–action pairs while ensuring numerical
stability through the upper bound.

To balance robustness, the penalty coefficient $\alpha$ is annealed during training using a linear decay schedule:
\begin{equation}
\alpha_t = \alpha_0
\left(
1-\frac{t}{T_{\max}}
\right),
\end{equation}
where $t$ denotes the training iteration index and
$T_{\max}$ is the total number of training steps.

Among the candidate uncertainty designs, we select the
count-based uncertainty penalty in our simulations after
empirical comparison with the other alternatives.
With this penalty, infrequently visited transitions incur
larger corrections to the Bellman target, suppressing
unreliable Q-value estimates and guiding the learned
policy toward well-supported regions of the dataset.
The robust advantage function is then defined as:
\begin{equation}
A^{\mathrm{robust}}(s,a)
=
Q_\theta(s,a)-V_\psi(s)-P(s,a),
\end{equation}

Finally, the policy is updated via advantage-weighted
behavior cloning:
\begin{equation}
L_\pi(\phi)
=
\mathbb{E}_{(s,a)\sim\mathcal{D}}
\Big[
\exp\!\left(\frac{A^{\mathrm{robust}}(s,a)}{\beta}\right)
\log \pi_\phi(a|s)
\Big].
\end{equation}
where $\beta$ is a hyperparameter controlling the strength of advantage weighting. By incorporating the uncertainty penalty into value estimation, Q-function regression, and policy optimization, DiSA-IQL effectively mitigates distribution shift and improves generalization for soft snake robot control.

\section{Simulation Results} \label{sec:exp}
In this section, we evaluate the DiSA-IQL framework for the target reaching of the soft snake robot on two representative benchmark tasks.

\begin{figure}[!ht]
    \centering
    \includegraphics[width=0.85\linewidth]{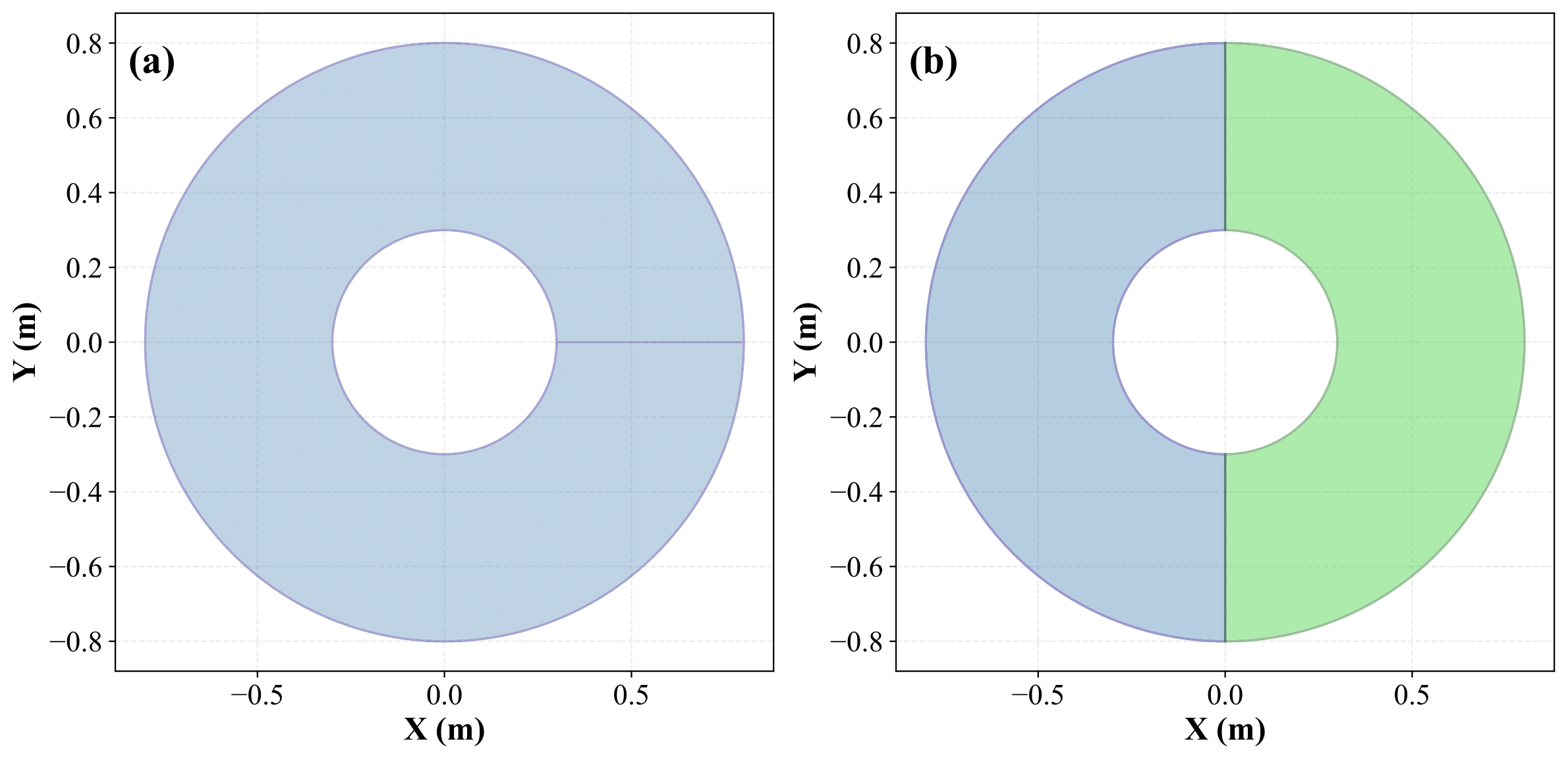}
    \caption{Illustration of goal-reaching tasks. 
    (a) In-distribution setting and 
    (b) OOD distribution shift setting.}
\label{fig:goal_regions}
\vspace{-15pt}
\end{figure}

\subsection{Simulation setup} 
The offline dataset was collected using a policy trained with a BER algorithm from our previous work \cite{Qi2024BER}. Specifically, we employed the best-performing checkpoint in the simulator to gather trajectories. Each trajectory contains observations (robot states), actions, rewards, next observations, terminal flags, and success indicators. The collected data were then formatted into a D4RL-style MDP dataset \cite{fu2020d4rl}, with both actions and rewards normalized to ensure stable training. To evaluate the effectiveness of the proposed approach, we compared DiSA-IQL with three representative offline RL algorithms: BC~\cite{pomerleau1988alvinn}, CQL~\cite{NEURIPS2020_0d2b2061}, and IQL~\cite{kostrikov2021offlinereinforcementlearningimplicit}.

To comprehensively assess performance and generalization, we evaluate the policies on two categories of tasks:
\begin{itemize}
    \item \textit{In-distribution setting}: both training and testing goals are sampled from the same region, serving as a baseline for performance in familiar environments (Fig.~\ref{fig:goal_regions}a).
    \item \textit{Out-of-distribution (OOD) setting}: policies trained in one half-annulus region are evaluated both within the training region and in the other half-annulus region, assessing transferability across structurally related but distinct goal spaces (Fig.~\ref{fig:goal_regions}b).
\end{itemize}

\subsection{In-distribution evaluation} 
\begin{figure}
    \centering
    \includegraphics[width=0.85\linewidth]{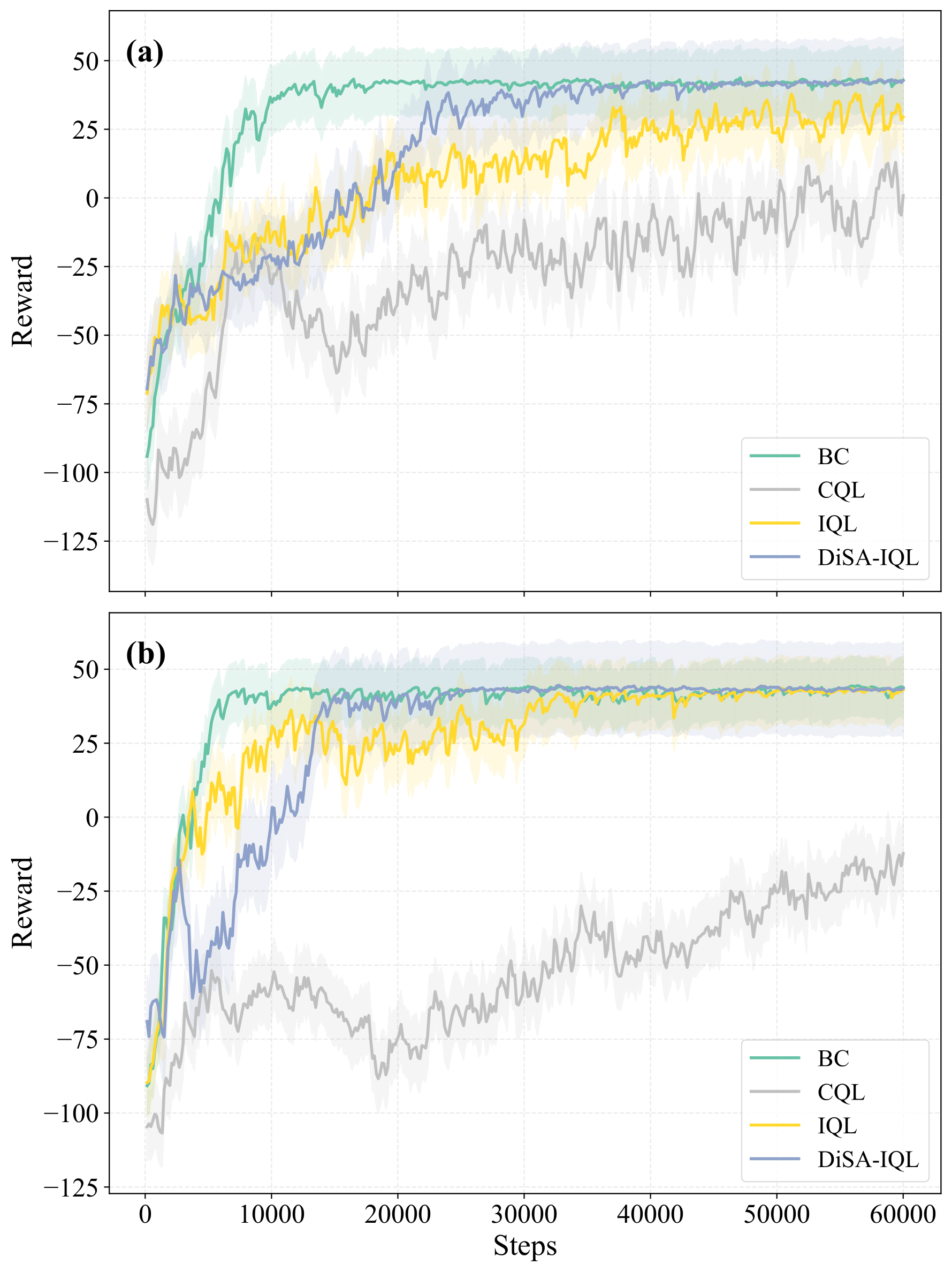}
   \caption{Training rewards in the (a) in-distribution setting and, (b) out-of-distribution setting.}
\label{fig:placeholder1}
\vspace{-14pt}
\end{figure}

In this section, we evaluate the performance of the proposed DiSA-IQL under the in-distribution setting (Fig.~\ref{fig:goal_regions}a), where the policies are both trained and tested in the same region. Fig.~\ref{fig:placeholder1} (a) presents the training reward curves comparing DiSA-IQL with three representative baselines (BC, CQL, and IQL) in the in-distribution setting. 
As shown in Fig.~\ref{fig:placeholder1} (a), all methods demonstrate stable learning under the in-distribution scenario. BC converges the fastest, primarily because it directly imitates the behavior policy without additional constraints. IQL and the proposed DiSA-IQL also converge quickly and reach a similar reward level as BC, with DiSA-IQL showing more stable performance. In contrast, CQL converges much more slowly and converges at a lower reward, indicating limited effectiveness in this setting. Additional parameter tuning may improve its performance.

Table~\ref{tab:metrics1} summarizes the quantitative evaluation results (500 testing points) under the in-distribution setting. DiSA-IQL achieves the highest performance, reaching a 100\% success rate with an average reward of 42.5 and an average of only 37 steps per episode. BC also performs competitively, achieving a 99.2\% success rate and the lowest average number of steps (36), though with slightly lower rewards compared to DiSA-IQL. IQL achieves a success rate of 93.3\%, but requires more steps on average (50), reflecting slower convergence during testing. In contrast, CQL lags significantly behind, with only a 65.8\% success rate, a much lower average reward (3.7), and a longer trajectory length (91 steps), indicating poor policy quality in this setting. 

\begin{table}[htbp]
\centering
\caption{Results of in-distribution of different algorithms.}
\begin{tabular}{lcccc}
\toprule
\textbf{Metrics} & \textbf{BC} & \textbf{CQL} & \textbf{IQL} & \textbf{DiSA-IQL} \\
\midrule
Success Rate & 99.2\% & 65.8\%  & 93.3\% & 100\% \\
Average  Reward   & 42.3  & 3.7  & 34.0  & 42.5 \\
Average  Steps     & 36 & 91     & 50      & 37 \\
\bottomrule
\end{tabular}
\label{tab:metrics1}
\vspace{-8pt}
\end{table}

\subsection{Out-of-distribution (OOD) evaluation} 
In this subsection, we evaluate the performance of the proposed DiSA-IQL under the out-of-distribution (OOD) setting (Fig.~\ref{fig:goal_regions}b). In this scenario, the policies are trained on targets located in the left region, while evaluation is conducted on both the same (training) region and unseen targets in the right region, simulating a distribution shift between training and testing.
Fig.~\ref{fig:placeholder1} (b) presents the training reward curves comparing DiSA-IQL with three representative baselines (BC, CQL, and IQL) in the out-of-distribution settings. The overall observations are similar to the in-distribution case, as training is performed on a collected dataset and most algorithms eventually converge to relatively stable rewards. However, CQL fails to converge satisfactorily and exhibits inferior performance compared to the other methods.

Table~\ref{tab:exp2} summarizes the quantitative evaluation results (500 testing points) under the out-of-distribution setting. DiSA-IQL achieves the best overall performance with a 91.2\% success rate, the highest average reward (27.9), and the fewest average steps (59), demonstrating strong generalization to unseen target regions. IQL ranks second, with a success rate of 75.8\% and an average reward of 17.1, but requires slightly more steps (76). BC performs moderately well, achieving 58.8\% success but with much lower rewards (6.5) and longer episodes (86 steps). CQL performs the worst, with only a 33.2\% success rate, a negative average reward, and the longest trajectories (118 steps), highlighting its inability to adapt under distribution shifts.
These results highlight that DiSA-IQL demonstrates superior robustness under distribution shift.

\begin{table}[htbp]
\centering
\caption{Results of out-of-distribution of different algorithms.}
\begin{tabular}{lcccc}
\toprule
\textbf{Metrics}  & \textbf{BC} & \textbf{CQL} & \textbf{IQL} & \textbf{DiSA-IQL} \\
\midrule
Success Rate & 58.8\% & 33.2\%  & 75.8\% & 91.2\% \\
Average Reward   & 6.5 & -22.5 & 17.1 & 27.9 \\
Average Steps     & 86  & 118    & 76     & 59     \\
\bottomrule
\end{tabular}
\label{tab:exp2}
\vspace{-8pt}
\end{table}

\begin{figure}
    \centering
    \includegraphics[width=1\linewidth]{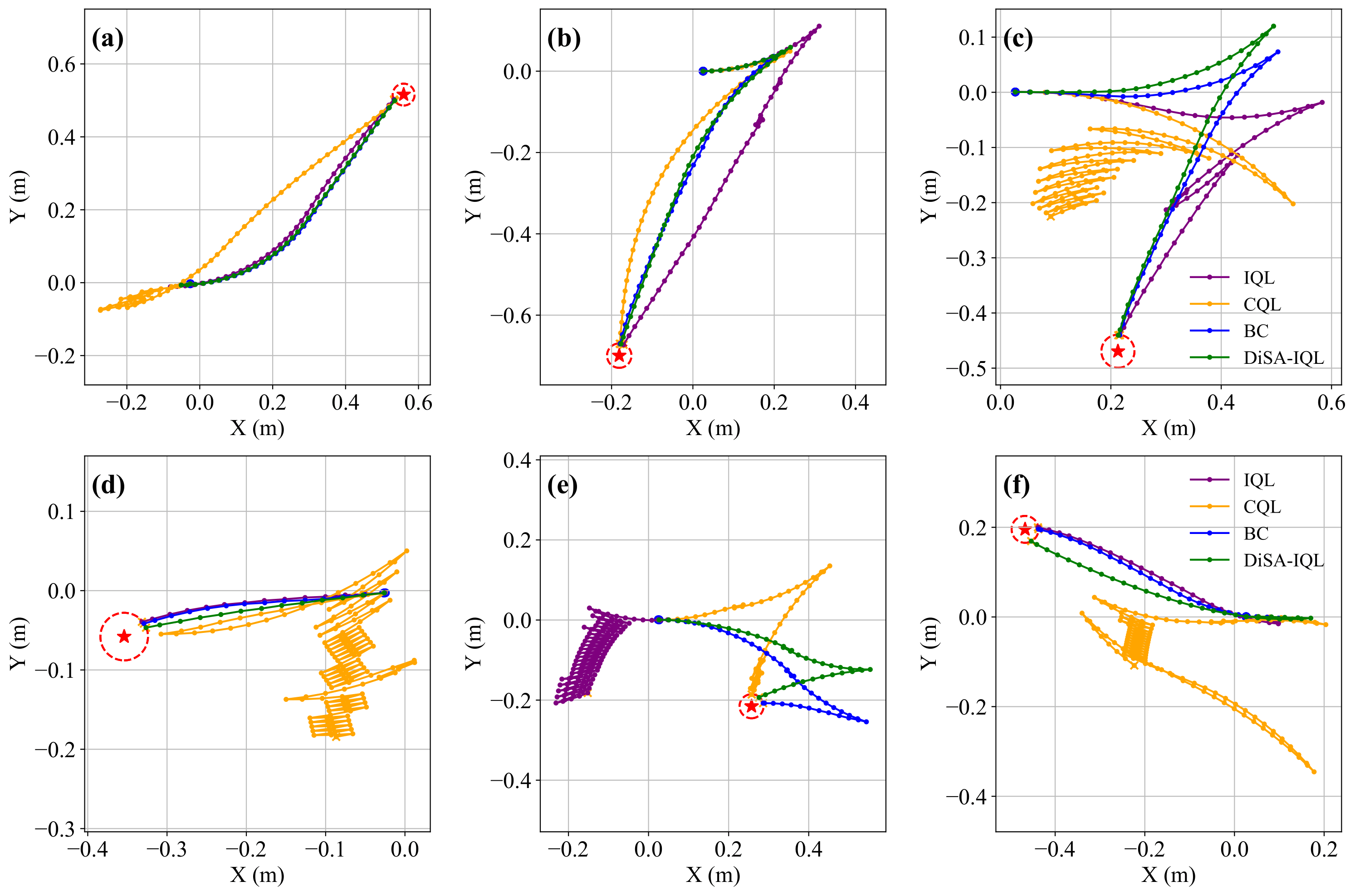}
    \caption{Trajectory performance of different algorithms (IQL, CQL, BC, and DiSA-IQL) across two evaluation scenarios. (a–c) In-distribution results. (d–f) Out-of-distribution generalization results. Red stars indicate target points, while red dashed circles denote the neighborhood region (radius = 0.03 m). Each row corresponds to a different evaluation setting.}
    \label{fig:placeholder7}
    \vspace{-14pt}
\end{figure}

Fig.~\ref{fig:placeholder7} visualizes representative trajectories of different algorithms across the three evaluation settings. In the \textit{in-distribution evaluation} (Figs.~\ref{fig:placeholder7}a–c), all methods except CQL are generally able to reach the targets, but the quality of trajectories differs. BC and DiSA-IQL produce relatively smooth and direct paths, while IQL tends to generate slightly more curved trajectories. CQL, on the other hand, exhibits occasional oscillations and deviations, suggesting instability even within the training distribution. In the \textit{out-of-distribution evaluation} (Figs.~\ref{fig:placeholder7}d–f), the differences among algorithms become more pronounced. BC often fails to generalize beyond the dataset, while CQL produces unstable trajectories due to excessive conservatism. IQL achieves moderate
generalization, whereas DiSA-IQL consistently produces smooth and reliable trajectories.

\vspace{-5pt}
\section{Conclusion}
This paper studies offline RL for soft snake robot control,
focusing on distribution shift and goal-space generalization. We constructed a large-scale offline dataset, benchmarked classical baselines, and proposed DiSA-IQL, an IQL extension with robustness modulation to suppress unreliable state–action pairs. Simulations showed that DiSA-IQL achieved stable learning in \textit{in-distribution} tasks, better generalization to \textit{out-of-distribution} scenarios, consistently outperforming Behavior Cloning, CQL, and vanilla IQL. While results confirmed the feasibility of offline RL for soft snake robots, this study is limited by its reliance on simulations, task-specific datasets, and sensitivity to extreme distribution shifts. Future work will validate DiSA-IQL on physical robots, expand datasets to more diverse terrains, and explore generative models and adaptive curriculum strategies to further enhance robustness.

\vspace{-5pt}
\section*{Acknowledgment}
The authors gratefully acknowledge Dr. Chuangchuang Sun at Villanova University for his insightful comments and constructive suggestions.

\vspace{-5pt}
\bibliography{ref}
\bibliographystyle{IEEEtran}

\end{document}